\documentclass[10pt, a4paper]{article}
\usepackage{lrec}
\usepackage{booktabs}
\usepackage{times}
\usepackage{latexsym}
\usepackage[textsize=tiny]{todonotes}
\usepackage{url}
\usepackage{tabularx}
\usepackage{multirow}
\usepackage{standalone}
\usepackage{blindtext}
\usepackage{hyperref}
\usepackage[utf8]{inputenc}
\usepackage[T2A,OT1]{fontenc}

\newcommand\textcyr[1]{{\fontencoding{T2A}\selectfont #1}}

\newcommand{\comment}[1]{}

\newenvironment{itemizerCompact}{\vspace{-1mm}
  \begin{itemize}
    \setlength{\itemsep}{2pt}
    \setlength{\parskip}{0pt}
    \setlength{\parsep}{0pt}
  }
{ \end{itemize}
  \vspace{-1mm}  }

\newcommand\surface[1]{{``#1''}}
\newcommand\lemma[1]{\underline{#1}}
\newcommand\translate[1]{{\emph{(#1)}}}
\newcommand\morph[1]{{\footnotesize \tt #1}}

\title{Neural disambiguation of lemma and part of speech \\
  in morphologically rich languages}

\name{
  Jos\'{e} Mar\'{\i}a Hoya Quecedo$^{\star}$,
  Maximilian W. Koppatz$^{\star}$,
  Giacomo Furlan$^{\star\dagger}$,
  Roman Yangarber$^{\star}$
}
\address{
  $^{\star}$University of Helsinki, Finland \\
  $^{\dagger}$University of Pisa, Italy \\
  {josemhoya@gmail.com, max.koppatz@gmail.com, g.furlan@studenti.unipi.it,
    roman.yangarber@helsinki.fi} \\
}

\abstract{
  We consider the problem of disambiguating the lemma and part of speech of ambiguous
  words in morphologically rich languages.
  We propose a method for disambiguating ambiguous words in context, using a large
  un-annotated corpus of text, and \comment{??? finite-state} a morphological
  analyser---with no manual disambiguation or data annotation.  \comment{??? We show that
    part-of-speech and lemma prediction disambiguate the majority of ambiguous
    analyses. }We assume that the morphological analyser produces multiple
  analyses for ambiguous words.
  The idea is to train recurrent neural networks 
  on the output that the morphological analyser produces for \emph{unambiguous} words.  We
  present performance on POS and lemma disambiguation that reaches or surpasses the state
  of the art---including supervised models---using {\em no manually annotated data}.
  We evaluate the method on several morphologically rich languages.  }

\begin{document}

\maketitleabstract{}

\section{Introduction}

The problem of disambiguation is defined as selecting the correct analysis from a set of
possible analyses for a word in a sentence---e.g., from among the analyses produced by a
morphological analyser. Disambiguation is performed by utilizing information in the
surrounding context.\footnote{This paper contains {\em corrigenda} to a previously
  published paper~\cite{2020-LREC-hoya-quecedo-disambig}.  It corrects a mistake in the
  original evaluation setup, and the results reported in Section~\ref{sec:exp}, in
  Tables~\ref{tab:meas}, \ref{tab:rez}, and~\ref{tab:err}.}

Morphological analysers are commonly used in various NLP applications. These normally produce a significant amount of ambiguous analyses. In this work we tackle the problem of disambiguation by training a model for predicting the correct part-of-speech (POS) and lemma. We show that for the majority of cases, this is sufficient to disambiguate from the set of possible analyses.

We use manually annotated data only for \emph{evaluation}, which means that to train our
model we need only a morphological analyser for the language and an unlabelled corpus.



The main idea of our approach is to use bidirectional LSTMs~\cite{gers2000}---BiLSTMs---to disambiguate the output of morphological analysers, by utilizing only the unambiguous outputs during the training procedure.
%
We train bidirectional models using a sequence of embeddings for the surface form for each target word. The objective of the network is to produce output probability distributions over the possible POS tags and lemmas. The model is trained using only the unambiguous input tokens; the loss is computed only for those unambiguous instances. Ambiguous tokens are not considered as target tokens during training.


Since we only input unlabelled data for training, the quality of the model itself is only
affected by the amount of available unlabelled data for the language. In our experiments,
we evaluate our models on manually annotated data sets for Finnish, Russian and
Spanish. For Finnish and Russian, at least, annotated (i.e., disambiguated) data is in
limited supply, whereas for all three languages unlabelled data is in abundant supply.


The paper is organized as follows.
In Section~\ref{sec:prior} we point to some relevant prior work.
In Section~\ref{sec:prob} we describe the problem of morphological ambiguity and provide a brief motivation for the interest in the problem.
In Section~\ref{sec:ambig} we provide a classification for the different types of ambiguity that appear in the corpus, as well as an analysis of the viable and appropriate strategies for each type of ambiguity.
Section~\ref{sec:model} describes our data pre-processing steps and model architecture.
Section~\ref{sec:exp} specifies our experimental setup, as well as the parameters used in training.
In Section~\ref{sec:res} we discuss the results obtained from the experiments.
Section~\ref{sec:conc} concludes with current directions of research.

\section{Related work}
\label{sec:prior}

There is an abundance of work on disambiguation in the context of various NLP tasks, we focus on just a few relevant ones here.

The work of~\newcite{yatbaz:2009} is conceptually similar to ours. Their work presents a
probabilistic model for selecting the correct analysis from a set of morphological
analyses for Turkish. Turkish and Finnish, as synthetic agglutinative languages, share the
problem of a high number of possible analyses for a given word. This limits the amount of
unambiguous data and presents a bigger problem than analytic or morphologically poor
synthetic languages such as English. The LSTM based approach by~\newcite{zalmout}, for
Arabic, is also similar to our method. They train a POS tagging model on an annotated
corpus, using added features, and use the resulting model to disambiguate a morphological
analyser, achieving a lemma accuracy of 96.8\%. The POS tagger
by~\newcite{inoue-etal-2017-joint} for Arabic utilizes a form of multi-task learning.
\newcite{DBLP:journals/corr/abs-1810-06908} present another neural morphological tagger,
for Estonian, in which the output of an analyser is also used to augment the input to
their neural models.

In contrast to the above mentioned neural models, we use the unambiguous outputs of the analyser to learn to disambiguate remaining ones, instead of learning a POS tagger on an annotated corpus.





\section{Problem description}
\label{sec:prob}

\subsection{Definitions}
Throughout this work, we make use of the following concepts:

\begin{itemizerCompact}

\item

  The \emph{part-of-speech} (POS) of a word (of a surface form) is its morpho-syntactic category or class. This indicates the role the word plays in the sentence, as well as the inflectional paradigm---the pattern of inflection---that the word follows. Examples of POS are: noun, verb, and adjective.\footnote{Many languages, including those we work with, distinguish \emph{open} vs.~\emph{closed} POS classes. In morphologically rich languages, open POSs are heavily inflected, whereas closed classes are not inflected, or have very limited inflection.}

\item

  The \emph{lemma} is the canonical, or ``dictionary,'' form of a word.  For example, for nouns the lemma is the nominative case singular and for verbs the lemma is the infinitive.\footnote{This is dependent on the language---the former holds true for Finnish and Russian, which are the languages with which we experiment in this paper, but not for other languages such as Latin.}

\item
  A \emph{surface form} is the form in which the word appears in text. The surface form may be an inflected form of the lemma, or may be identical to the lemma; for uninflected POSs, the surface form is always identical to the lemma.

\item

  \emph{Morphological tags} are values that the morphological analyser assigns to morphological features of the word. For example, the feature \emph{number} may have values such as \emph{singular} and \emph{plural}; the feature \emph{case} may have values such as \emph{nominative} and \emph{genitive}, depending on the feature inventory of the language.
  
\end{itemizerCompact}

Morphological analysis is the task of breaking down a surface form into its lemma, POS and morphological features (tags), by means of a morphological analyser.
As an example, consider the Finnish surface form ``kotiin'' (into/toward home).
A morphological analysis of ``kotiin'' would be:
\begin{center}
  \texttt{koti+N+Sg+Ill}
\end{center}
This indicates that the lemma is \texttt{koti}, the POS is \texttt{N} (Noun), and the morphological features are \texttt{Sg} (singular number) and \texttt{Ill} (illative case, meaning ``into/toward'').

\subsection{Ambiguity}
\label{sec:ambiguity}

Natural language is inherently ambiguous, and there are many ways in which ambiguity manifests itself. For written text, we have several types of ambiguity.
\emph{POS} ambiguity is a kind of syntactic ambiguity, where a word may be considered to have one of several syntactic roles inside a sentence.
\emph{Lemma} ambiguity occurs when a surface form is a form of more than one lemma.
\emph{Morphological} ambiguity occurs when a surface form has several possible
analyses---several sets of morphological tags.
\emph{Word sense} ambiguity---when a single lemma may have several different meanings.

In spoken language, other kinds of ambiguities exist, such as homophones---two words which are written differently but are pronounced the same. Spoken language ambiguity is outside the scope of our work, we concentrate on written text.

One example of ambiguity is the Finnish surface form \surface{tuli}, which has the following analyses:\\
\lemma{tuli} \translate{fire} \morph{Noun, nominative, sing.} $||$ \\
\lemma{tulla} \translate{come} \morph{Verb, indicative, active,\\
  \indent past, 3rd person, sing.}  \hspace{1cm}

This exhibits all of the above kinds of ambiguity: POS, lemma, morphological tags, and
word sense are all ambiguous.

Disambiguation is a central problem in many NLP tasks, for many reasons.
Morphological disambiguation in morphologically rich languages is crucial for translation.
In our application setup~\cite{katinskaia:2018-lrec:revita,katinskaia:2018-DHN:revita}, we
build tools to aid in language learning,\footnote{\url{revita.cs.helsinki.fi}}
When a student points at an unfamiliar surface form in the text, which happens to be
ambiguous, we need to identify the correct lemma \emph{appropriate to the context}---so as
not to confuse the learner with extraneous translations.\footnote{In other words: in our
  work, we are not concerned with word-sense ambiguity \emph{alone}---only in conjunction
  with POS ambiguity or lemma ambiguity.}
Especially in morphologically rich languages such as Finnish and Russian, unambiguous
lemmatization is central for NLP applications that build a vocabulary from corpora.  For
these languages the size of the vocabulary becomes very large without lemmatization.  If
the lemmatization is ambiguous, then subsequent models are based on an inaccurate
vocabulary.


Our approach is based on the assumption that the context \emph{primes} the selection of the appropriate reading from a set of several readings for an ambiguous surface form. By ``priming,'' we mean the following: a simple experiment with Google's translator shows that the ambiguous word \textcyr{белки}, is easily disambiguated by its immediate context.  The surface form has two lemmas: ``\textcyr{белка}'' (squirrel) and ``\textcyr{белок}'' (protein).  Google easily translates ``\textcyr{белки и медведи}'' as ``squirrel and bears'', whereas it translates ``\textcyr{белки и углероды}'' into ``proteins and carbons.''

Thus, Google's translation problem subsumes the disambiguation problem that we are trying to solve; in fact, Google's translator could be viewed as a ``poor man's solution'' to the disambiguation problem.
However, because we are trying to solve the simpler problem---disambiguation---key point is that we may be able to solve it with a more lightweight solution. This would offer 3 benefits:
A. we could achieve it with fewer and cheaper resources---translation requires supervision from massive parallel corpora;
B. we may be able to achieve it with simpler models;
and C. we may be able to achieve better performance on disambiguation, than if we tried to use a full translation machine to perform disambiguation.
We further rely on the assumption that a large corpus will contain enough unambiguous contexts for each POS and for each lemma, so that the model should be able to learn to disambiguate the ambiguous instances.

\section{Types of ambiguity}
\label{sec:ambig}

We discuss briefly a taxonomy of the types of ambiguity that are of interest to us.
Additional examples are given in Appendix~\ref{app:examples}, for several languages.
In many cases, the problem of disambiguation can be reduced to one of two problems: POS
tagging or lemmatization---given a surface form in context (running text), find its POS or
lemma, respectively.
We outline the main types of morphological ambiguity, and whether we can use one approach
or the other to resolve it.

We classify lemmas into two types---depending on whether they accept inflectional
morphemes: \emph{declinable} lemmas accept them, and \emph{indeclinable} lemmas do not.  Thus,
an indeclinable lemma has only one surface form.  Declinable lemmas can have many surface
forms.

We use the term \emph{reading} to denote a unique combination of lemma and POS\@.

We divide surface form ambiguities into three categories in the following subsections: two
(or more) declinable lemmas, one declinable and one indeclinable lemma, or two indeclinable
lemmas.

\subsection{Surface forms with two declinable lemmas}

This is the easiest case to train for, since, in general, the sets of surface forms
derived from the two lemmas rarely overlap.
For example, Finnish surface form FI \surface{tuli} has two readings, as above:\\
\lemma{tuli} \translate{fire} \morph{Noun, nominative, sing.} $||$ \\
\lemma{tulla} \translate{come} \morph{Verb, indicative, active,\\
  \indent past, 3rd person, sing.}  \hspace{1cm}

In this example, the lemmas and the POS's of the readings are both different.  This is the
most common type of ambiguity, and either method (POS or lemma disambiguation) can be
applied.

If the lemmas of the readings are identical, we cannot use lemmatization to resolve the
ambiguity, and must resort to POS disambiguation, e.g.:\\
RU \surface{\textcyr{знать}}: \translate{know} \morph{Verb} $||$ \translate{nobility}
\morph{Noun}

Conversely, if the POS's of the readings are the same, but the lemmas are
different, we cannot use POS tagging to resolve the ambiguity:

When two readings are the same for a surface form---i.e., the lemma and POS are the same,
but the morphological tags are different---our methods are not suitable to disambiguate:
e.g.,
FI \surface{nostaa}{:} \\
\lemma{nostaa} \morph{Verb, infinitive} $||$ \\
\lemma{nostaa} \morph{Verb, present, indic., 3rd, sing.}

\begin{table}
  \def\tabularxcolumn#1{m{#1}}
  \begin{tabularx}{\columnwidth}{XXXX}
    \toprule
    & Declinable-\newline Declinable & Declinable-Indeclinable  & Indeclinable-Indeclinable\\
    \midrule
    \( \neq \) POS \newline \( = \) lemma & POS disamb. & POS disamb. & POS disamb. \\
    \midrule
    \( = \) POS \newline \( \neq \) lemma & lemma disamb. & n/a & n/a \\
    \midrule
    \( \neq \) POS \newline \( \neq \) lemma & either & POS disamb. & n/a \\
    \midrule
    \( = \) POS \newline \( = \) lemma & neither & n/a & n/a \\
    \bottomrule
  \end{tabularx}
  \caption{Viable approaches for each type of ambiguity.}
  \label{tab:sum-ta}
\end{table}


Lastly, we turn to word-sense ambiguity.
For example, in English, the word/lemma ``spirit'' may mean ``soul'' or ``alcohol''.
These are unrelated semantically, but have the same lemma, same POS, and follow identical
inflectional patterns.  Although this type of ambiguity is also important for translation,
it is outside the scope of this paper.\footnote{However, we are interested in
  distinguishing the Noun ``spirit'' from the Verb ``spirit'', which has a different
  meaning---``to move away briskly or secretly''---but also crucially has a different
  POS.}

To sum up, we are concerned with disambiguating among different \emph{readings}---i.e., POS
or lemma disambiguation.  We are not concerned with disambiguating different possible
morphological tags of a given reading, nor with disambiguating multiple word senses of a
given reading.

\subsection{Surface forms with one declinable and one indeclinable lemma}
\label{sec:1var+1invar}

In this case, trying to predict the lemma may be less effective:
although the lemmas may be different, \emph{every} instance of the reading with the
indeclinable lemma is ambiguous---since it always ``drags along'' the other readings with
it.

Finnish and Russian have many instances of such surface forms.
In Finnish, many adverbs or post-positions originate historically from an inflected form
of a semantically related noun.
For example, FI \surface{j\"{a}lkeen}:\footnote{The post-position ``j\"{a}lkeen'' (after)
  is an ossified form of an inflection of the noun ``j\"{a}lki'' (footprint)---in the
  sense of ``after'' meaning ``in the footsteps of''. However, most analysers will analyse
  the post-position ``j\"{a}lkeen''
  as a separate lemma, not explicitly linked to its old nominal origin.}\\
\lemma{j\"{a}lkeen} \translate{after}  \morph{Post-position} $||$ \\
\lemma{j\"{a}lki} \translate{into a footprint} \morph{Noun, illative, sing.}\\
Thus, every occurrence of the post-position ``j\"{a}lkeen'' drags along with it the
readings for the illative case of ``j\"{a}lki'' (which is also a valid reading of
``j\"{a}lkeen'').

However, the model can still hope to learn that the POS of this surface form is post-position, since other \emph{unambiguous} post-positions may occur in similar contexts elsewhere in the corpus.
Thus, POS tagging is an effective solution to this type of ambiguity.

The POS determines whether the reading is declinable or indeclinable. Thus a surface form
cannot have a declinable and an indeclinable reading with the same POS, as seen in
Table~\ref{tab:inc-ta}, and in fact such instances do not appear in the corpus.

\begin{table}
  \def\tabularxcolumn#1{m{#1}}
  \centering
  \begin{tabularx}{\columnwidth}{XXXX}
    \toprule
    & Declinable-\newline Declinable & Declinable-Indeclinable & Indeclinable-Indeclinable\\
    \midrule
    \( \neq \) POS \newline \( = \) lemma & 8.78\% & 1.63\% & 6.47\% \\
    \midrule
    \( = \) POS \newline \( \neq \) lemma & 8.93\% & 0.00\% & 0.00\% \\
    \midrule
    \( \neq \) POS \newline \( \neq \) lemma & 40.29\% & 27.88\% & 0.00\% \\
    \midrule
    \( = \) POS \newline  \( = \) lemma & 6.04\% & 0.00\% & 0.00\% \\
    \bottomrule
  \end{tabularx}
  \caption{Incidence of each type of ambiguity in the Finnish corpus.}
  \label{tab:inc-ta}
\end{table}


\subsection{Two indeclinable lemmas}

If the readings are both indeclinable---neither can be inflected---and if their lemmas are
different, there is no ambiguity, as the surface forms will always differ.  If the lemma
and POS are the same, the readings must be trivially identical, since there are no other
morphological features.
If the POS is different, we can disambiguate via POS tagging, as in the previous
category,~\ref{sec:1var+1invar}.
It is important to note that, since these readings \emph{always} go together, in our work
we consider these ambiguities to be ``uninteresting'', since they can be considered
extra-morphological.\footnote{As is the case for English ``around,'' which can be either
  an adverb or a preposition.}


Table~\ref{tab:sum-ta} provides a summary of the effective approaches for each category.

In table~\ref{tab:inc-ta}, we see that POS tagging can effectively disambiguate most
cases, except those in which the POS is the same across readings.

\section{Model}
\label{sec:model}

We next turn to the technical description of our approach.
First, we outline the steps involved in preparing the data for our model. We then proceed
to present the architecture of our model, and the training procedure.

\begin{figure}[t]
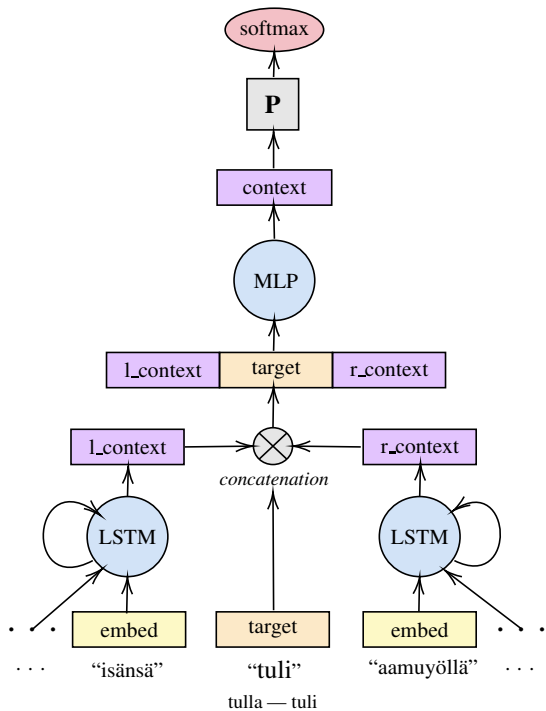

  \includestandalone[width=7.5cm]{network}
  \caption{Overview of the network (see Section~\ref{sec:architecture}).}
  \label{fig:network}
\end{figure}

\subsection{Data pre-processing}

First, we tokenize each document as a flat list of surface forms (tokens).

We then use morphological analysers to obtain the readings of each surface form.
For Finnish, we use analysers from the Giellatekno
platform~\cite{moshagen:2013:building-giellatekno}\footnote{\url{giellatekno.no}}. Giellatekno
analysers are based on Two-level Morphology, by~\newcite{koskenniemi:1983-thesis}.
For Russian, we use the analyser from~\newcite{klyshinsky2011method}.
For Spanish, we use the analyser from~\newcite{forcada2011afp}.

Since the goal is to \emph{disambiguate} the output of the analyser, the coverage of said analyser---the percentage of tokens that have an analysis---is a relevant concern. The Finnish, Russian and Spanish analysers have 95.14\%, 97.79\% and 96.78\% coverage, respectively. Most of the unknown tokens are foreign or misspelled words.

For Finnish---which has compounding---we split the surface form of the compounds into
their ``maximal'' pieces, i.e., the largest parts for which there is a lemma in the
analyser's lexicon.
For example, the Finnish compound word \textit{el\"{a}inl\"{a}\"{a}k\"{a}riasema} (``veterinary clinic'') is made up of three elementary stems: \textit{el\"{a}in} (``animal'') + \textit{l\"{a}\"{a}k\"{a}ri} (``doctor'') + \textit{asema} (``station'').
However, since the analyser has \textit{el\"{a}inl\"{a}\"{a}k\"{a}ri} (``veterinarian'') in its lexicon, we split as \textit{el\"{a}inl\"{a}\"{a}k\"{a}ri} + \textit{asema}. This helps us keep the vocabulary smaller---since there is a potentially infinite number of possible compounds in Finnish---while keeping the meaning of commonly used compounds, which usually differs a little from that of the sum of its parts.

For Russian, this is not a concern, as there are generally no compound words. There may be cases in which a lemma is formed by joining two other lemmas, but this is considered a new lemma in its own right. The same applies to Spanish, where we additionally have clitic pronouns attached to verbs and prepositional contractions (preposition + article), which are treated as separate tokens.

While we do preserve information about sentence boundaries in the form of punctuation, we do not explicitly preserve sentence structure in terms of the training window. We found that several sentences in our corpora were too short to provide the contextual information necessary for disambiguating the target words, and that this information was partially found in the adjacent sentences. Instead, we make a sliding window of radius $r$ over this list of tokens, i.e.\ we take $r$ tokens to the left and $r$ tokens to the right of some given target token, as well as the token itself.

Tokens are selected as targets for the training set only if they are unambiguous. Each training instance consists of said window, and the label for the target word, given by the analyser---the lemma or POS, depending on the desired target for the model. The target for the lemma is the index of said lemma in our vocabulary.

For the test set, we instead select only the ambiguous tokens, since the unambiguous ones will trivially give us a 100\% accuracy. Each testing instance consists of the window, the possible labels and the true label for the target word.

\begin{table}
  \centering
  \begin{tabular}{lr}
    \toprule
    Window size & 21\\
    Embedding size & 300\\
    LSTM hidden units & 512\\
    LSTM layers & 1\\
    MLP hidden neurons & 1024\\
    \bottomrule
  \end{tabular}
  \caption{Settings for the network.}
  \label{tab:mpar}
\end{table}


We then obtain the word embeddings for each surface form in the window using the FastText~\cite{FastText} Common Crawl\footnote{\url{https://commoncrawl.org/}} pre-trained models
for each language. This allows us to get an embedding even for out-of-vocabulary words,
and to efficiently get embeddings for Finnish, which has a very large surface form
vocabulary in our corpus---around 2 million unique tokens. FastText allows us to avoid
using an embedding matrix and instead obtain the embeddings dynamically during training.

For the positions in the window which go beyond the limits of the document, we insert a zero-valued embedding as padding.

Each language has its own morphological analyser, and their outputs differ slightly.  For
compatibility between different analysers and our model, we map all POS to a common
universal set. This also allows us to simplify the problem, by aggregating POS that fulfil
a similar role, such as: postpositions + prepositions $\rightarrow$ adpositions.

We use a set of 10 POS\@: noun, pronoun, numeral, adjective, verb, adverb, adposition, conjunction, punctuation, other (a catch-all category for things like acronyms or symbols).

Lemmas which are composed entirely of numerical digits are assigned a special embedding for numbers, since we consider them to always have a more or less equivalent role in the context.

Ambiguities between common nouns and proper nouns are ignored, as names are out of the scope of what we try to accomplish here and should be solved using Named Entity Recognition (NER) techniques instead.

We keep punctuation in order to recover the sentence boundaries within the window, as
these may still prove useful to generate a proper context.

\begin{table}
  \centering
  \begin{tabular}{lr}
    \toprule
    Batch size & 50\\
    Dropout rate & 0.1\\
    Adam \( \beta_1 \) & 0.9\\
    Adam \( \beta_2 \) & 0.999\\
    Adam \( \alpha \) & 0.0001\\
    Adam \( \epsilon \) & 0.001\\
    Epochs & 20 \\
    \bottomrule
  \end{tabular}
  \caption{Settings for the training hyper-parameters.}
  \label{tab:thyp}
\end{table}


\comment{
  GIACOMO: T5
  
75.90	73.50	73.90
61.06	66.38	61.86

80.50	79.60	79.60
60.62	65.90	61.38

81.63	78.33	79.07
70.78	69.15	67.50

T6:

64.80	73.50	97.79	97.70
31.10	66.10	97.68	95.30

75.22	79.60	98.47	96.90
12.33	69.33	96.55	81.70

76.20	79.87	96.92	89.00
12.33	69.46	95.27	88.00

T6:

70.23  &  74.77  &  78.77  &  77.10  &  61.47 \\

82.60  &  75.65  &  79.80  &  78.60  &  78.15 \\

86.80  &  70.75  &  58.83  &  79.82  &  76.00 \\

}

\subsection{Architecture \& training}
\label{sec:architecture}

\begin{table*}
  \centering
  \begin{tabular}{ llrrrr }
    \toprule
    Language & Target & \% Ambiguous & Precision & Recall & \( F_1 \) score \\
    \cmidrule{1-6}
    \multirow{2}{*}{Finnish}
             & POS   & 8.9  & 75.90 & 73.50 & 73.90 \\
             & Lemma & 8.8  & 61.06 & 66.38 & 61.86 \\
    \cmidrule{1-6}                                 
    \multirow{2}{*}{Russian}                       
             & POS   & 7.7  & 80.50 & 79.60 & 79.60 \\
             & Lemma & 10.2 & 60.62 & 65.90 & 61.38 \\
    \cmidrule{1-6}                                 
    \multirow{2}{*}{Spanish}                       
             & POS   & 15.3 & 81.63 & 78.33 & 79.07 \\
             & Lemma & 15.5 & 70.78 & 69.15 & 67.50 \\

    \comment{
             & POS   & 8.9  & 0.81 & 0.79 & 0.79 \\
             & Lemma & 8.8  & 0.80 & 0.75 & 0.76 \\
    \cmidrule{1-6}
    \multirow{2}{*}{Russian}
             & POS   & 7.7  & 0.85 & 0.84 & 0.84 \\
             & Lemma & 10.2 & 0.82 & 0.77 & 0.80 \\
    \cmidrule{1-6}
    \multirow{2}{*}{Spanish}
             & POS   & 15.3 & 0.83 & 0.81 & 0.81 \\
             & Lemma & 15.5 & 0.82 & 0.75 & 0.75 \\
    }
    \bottomrule
  \end{tabular}
  \caption{Evaluation results for each model. The column \emph{\% Ambiguous} shows the
    percentage of ambiguous tokens in each corpus.}
  \label{tab:meas}
\end{table*}

\begin{table*}
  \centering
  \begin{tabular}{ l|lrr|rrl } 
    \toprule
    Language & Target & Blind & Guided & Token & SOTA & \\
    \cmidrule{1-7}
    \multirow{2}{*}{Finnish}
             & POS           & 64.80 & 73.50 & {\bf 98.1} & 97.70 & \cite{udst:turkunlp} \\
             & Lemma         & 31.10 & 66.10 & {\bf 97.9} & 95.54 & \cite{udst:turkunlp} \\
    \cmidrule{1-7}                                         
    \multirow{2}{*}{Russian}                               
             & POS           & 75.22 & 79.60 & {\bf 98.6} & 96.90 & \cite{dereza2016automatic-SOA-RUSSIAN} \\
             & Lemma         & 12.33 & 69.33 & 98.3 & {\bf 98.99} & \cite{2017-kotelnikov-rus-morpho} \\ 
    \cmidrule{1-7}                                         
    \multirow{2}{*}{Spanish}                               
             & POS           & 76.20 & 79.87 & {\bf 97.0} & 89.00 & \cite{2015-escartin-spanish} \\
             & Lemma         & 12.33 & 69.46 & {\bf 96.1} & 88.00 & \cite{2015-escartin-spanish}\\
    \bottomrule
  \end{tabular}
  \caption{The columns mean:
    \emph{Blind}: percentage of ambiguities resolved with ``blind'' predictions---without
    using the analyser output.
    \emph{Guided}: percentage resolved by picking the highest-scoring prediction from the
    analyser output.
    \emph{Token}: overall \emph{token-level} accuracy, by applying the best
    method.
    \emph{SOTA}: for comparison, shows the state-of-the-art results.  }
  \label{tab:rez}
\end{table*}


\begin{table*}
  \centering
  \begin{tabular}{ lrrrrr }
    \toprule
    Language & Noun & Adjective & Verb & Adverb & Other\\
    \midrule
    Finnish & 70.23  &  74.77  &  78.77  &  77.10  &  61.47 \\
    Russian & 82.60  &  75.65  &  79.80  &  78.60  &  78.15 \\
    Spanish & 86.80  &  70.75  &  58.83  &  79.82  &  76.00 \\

    \comment{
    Finnish & 75.8 & 82.8 & 78.3 & 74.7 & 75.5 \\
    Russian & 77.4 & 79.8 & 90.7 & 82.6 & 92.5 \\
    Spanish & 82.6 & 71.6 & 86.3 & 86.6 & 82.7 \\
    }

    \bottomrule
  \end{tabular}
  \caption{Accuracy (percent) for each POS.}
  \label{tab:err}
\end{table*}


Our model architecture is based on context2vec~\cite{context2vec}, which itself is a modification of the original word2vec CBOW model~\cite{word2vec}. In context2vec, the context for each word is computed using a bidirectional LSTM, rather than as a vector average (as in word2vec), which enables the embeddings to capture sentence structure and word order, rather than only word co-occurrence.

The training procedure is analogous to that of word2vec, since our objective is similar: given a context, compute the probability that the word belongs to that context, for each word in the vocabulary---in our case, instead of word tokens, we compute the probability for the lemma or the POS\@. Thus, our input and target vocabularies are different, in contrast to word2vec.

The architecture borrows some ideas from neural machine translation (NMT) encoder-decoder models, such as that developed by Google~\cite{gtrans}. In that model, encoding the context of a token into one vector is enough to be able to translate---and therefore disambiguate---that token. We therefore use the encoder part of the architecture to capture the necessary information to disambiguate a token.

The model consists of three trainable parts:
\begin{itemizerCompact}
\item
  Bi-LSTM which produces the left and right context embeddings.
\item
  multi-layer Perceptron (MLP), which merges these into a single context embedding.
\item projection matrix, to transform the context embedding into scores for all possible
  labels.
\end{itemizerCompact}

Each training instance consists of a window of surface form embeddings around an unambiguous target word, and the label for such word, defined as the index of the corresponding lemma or POS in the vocabulary.

To obtain the predictions, we proceed as follows.
First, we feed the window from the beginning to the target word to the left LSTM, and from the end to the target word to the right LSTM\@. Their hidden states serve as the left and right context embeddings.

We then concatenate the left and right context with the surface form embedding for the target word, and feed the result to the MLP\@. This ``residual'' connection, where some input is fed to several layers of the network, is also a concept from NMT, and is done in order to separate important parts of the input from the encoded state.

Next, the output of the MLP is multiplied by the projection matrix, to get an array of scores of length equal to the number of possible labels.

Finally, we apply a softmax function to get the probability distribution for all labels. An overview of the model can be observed in Figure~\ref{fig:network}.\footnote{The input (bottom line) is a window of words in Finnish: ``\dots is\"ans\"a tuli aamuy\"oll\"a \dots'' (``\dots his father arrived in the early hours \dots'')  The \emph{target} surface form, for which we will try to get a prediction, is ambiguous: it may be a verb (arrive) with lemma ``tulla'', or a noun (fire) with lemma ``tuli.'' (This is the same as the example discussed in Section~\ref{sec:ambiguity}.) }

To obtain the loss for the model, we compute the cross-entropy between the predicted probability distribution and the real distribution, which is the one-hot encoding of the true label index.

\section{Experiments}
\label{sec:exp}

\subsection{Data}

The data we use are obtained from two different sources.  The annotated evaluation data is
from the Universal Dependencies Treebank~\cite{udtreebank}.  These data are in the CoNLL-U
2006/2007 format~\cite{conll2007}.  The annotations in the data are used for determining
the stopping criteria and for evaluation of the resulting models.  For Finnish, as the
annotated data sets were quite small, we used an unlabeled collection of 600K proprietary
news articles for training the model, after processing the text with the Finnish
morphological analyser.  The Russian annotated data set was large enough to use its
predefined train-test split.
%

\subsection{Experimental setup}

For each language, we trained two separate models: one to predict the correct POS, and one
to predict the lemma.  We train the model on the unambiguous analysed tokens; we do not
train on the ambiguous instances---this allows us to explore the unsupervised approach,
with no need for manual disambiguation. 

In each case, we evaluate our models by two metrics. First, we pick the analysis with the
highest value in the softmax output probability vector.  We call this the ``blind''
disambiguation.  Secondly, rather than picking the highest-scoring softmax output overall,
we pick the highest-scoring output from the output probability vector, but choose only
from among the options deemed possible by the morphological analyser.  This is the
``guided'' approach.
For example, if the model is predicting POS, the blind approach selects the POS that
receives the highest output score from all possible POSs in the language. The ``guided''
(analyser-based) approach selects the highest scoring POS only from those POS values that
are among the possibilities admitted by the morphological analyser for the given surface
form.
We proceed analogously for the lemma-based models.

The ``blind'' predictions are thus equivalent to plain POS tagging and lemmatization.

We evaluate each model with the manually annotated, disambiguated corpus for each language. We compute both evaluation metrics (precision, recall and \( F_1 \) score) as well as the percentage of correct predictions, for direct comparison with the state of the art.

In addition, we evaluate each metric with respect to a ``confidence'' measure, defined as the probability given by the softmax function for each prediction. To this end, we set a confidence threshold \( \theta_{conf} \) such that any prediction with confidence below that threshold will be deemed invalid. In doing so, we wish to test whether the more confident predictions will have a higher precision without a significant loss in recall, for applications where the goal is to obtain the highest possible precision.

Table~\ref{tab:mpar} details the parameters used for the network.  Increasing the number
of trainable parameters yields no significant increase in accuracy\comment{??? , but we
  are constrained by hardware limitations}.  It is possible that with a more complex model
the prediction accuracy could be slightly higher.

Table~\ref{tab:thyp} details the training hyper-parameters. We use the Adam optimizer,~\cite{adam}, to minimize the loss.

For each language, we first split the small data with manually resolved ambiguities into a
{\em development} set (10\%) and an {\em evaluation} set (90\%).
We repeat the experiment 10 times, each time with a random development/evaluation split,
and different random seeds.
The development set is used to determine the {\em stopping criteria}---when to stop
training on the unlabeled data.
All reported results have been averaged over 10 random repetitions.


\section{Results}

\label{sec:res}
\begin{figure}
  \resizebox{\columnwidth}{!}{\input{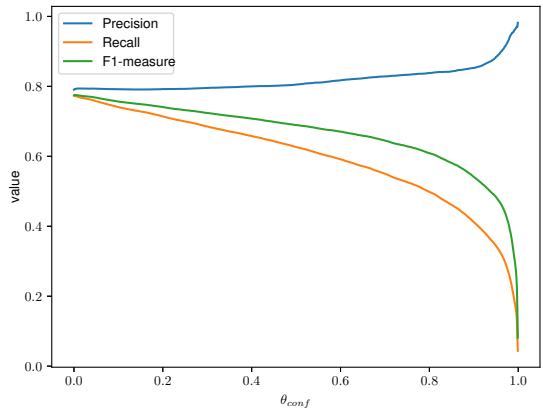}}
  \caption{Finnish POS confidence vs.\ metrics.}
  \label{fig:fipos}
\end{figure}

\begin{figure}
  \resizebox{\columnwidth}{!}{\input{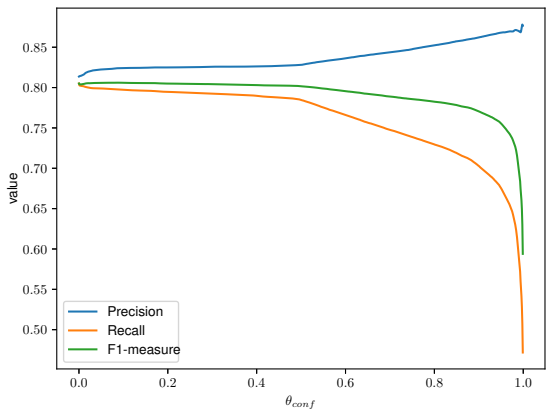}}
  \caption{Russian POS confidence vs.\ metrics.}
  \label{fig:rupos}
\end{figure}

\begin{figure}
  \resizebox{\columnwidth}{!}{\input{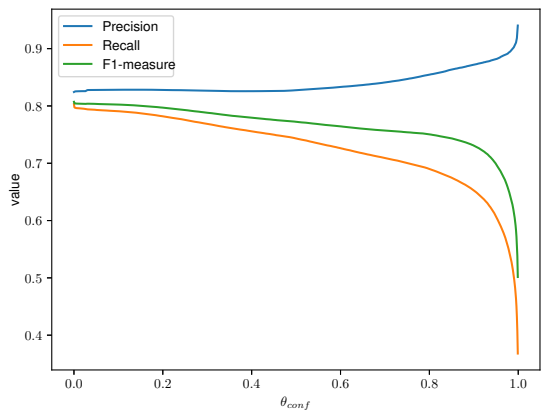}}
  \caption{Spanish POS confidence vs.\ metrics.}
  \label{fig:espos}
\end{figure}

For the three languages on which we performed an evaluation of our models, we significantly reduced the number of remaining ambiguities. Table~\ref{tab:meas} illustrates the results of our experiments in terms of number of ambiguities and evaluation metrics. Table~\ref{tab:rez} shows a comparison between our results and the state of the art.

While the Finnish and Russian analysers are much less ambiguous than the Spanish one, our
model is able to disambiguate the Spanish output to very nearly the same token-level
accuracy. Thus, our method is not reliant on a low percentage of ambiguity to begin with,
but instead other factors---such as the overlap in surface forms for a given pair of
lemmas---are much more relevant.


%
For Russian, the best result to date for POS tagging was reported
by~\newcite{dereza2016automatic-SOA-RUSSIAN}, achieved using
TreeTagger,~\cite{2013:Schmid-probabilisticpart-of-speech-treetagger}, at 96.94\%. We
could not find lemmatization results for Russian, but the work by~\newcite{pymorphy}
solves the broader problem of morphological ambiguity with an accuracy of 81.7\%.

For Finnish POS tagging and lemmatization, the TurkuNLP neural model~\cite{udst:turkunlp} achieves 97.7\% and 95.3\% accuracy, respectively, evaluated on the same dataset as our method.

For Spanish POS tagging and lemmatization, the model by~\newcite{carreras04} achieves an
accuracy of 89\% and 88\%, respectively, according to the evaluation done
by~\newcite{2015-escartin-spanish}.

As for the confidence analysis, we see that, for every language, we can in fact build a POS model which has very high precision (\textgreater 0.9) at the cost of being unable to obtain a prediction for a fraction of the instances. Figure~\ref{fig:fipos}, Figure~\ref{fig:rupos} and Figure~\ref{fig:espos} show the results for Finnish, Russian and Spanish, respectively.

\section{Conclusions}
\label{sec:conc}
We have shown that the output of morphological analysers can be disambiguated to a significant degree for Finnish, Russian and Spanish. The requirements for this procedure are: the language must have a morphological analyser, there must exist a text corpus, and preferably a small amount of annotated data for evaluation purposes. The same procedure we used should perform comparatively for any language with a morphological analyser, assuming it is of sufficient quality---unknown tokens must rely on the less accurate ``blind'' predictions for inference. There are many morphologically rich languages that could benefit from this, such as other Uralic languages, Turkic languages, many Indo-European languages, etc. There is limited annotated training data for many of these languages, but morphological analysers are available for most of them.

The quality of the analyser in terms of percentage of unambiguous output does affect the final total token accuracy. The difference between the two cases end result presented in this work was small in the end. It is unclear how much ambiguity will begin to significantly impair our method.

Named Entity Recognition (NER) could theoretically be used in conjunction with our procedure to further disambiguate the proper noun analyses.

We have achieved different performance depending on whether the objective used was disambiguating the lemma or POS\@. We have seen that different types of ambiguity are solved to varying degrees by predicting either POS or lemma. A natural next step would be to combine the two different models in an ensemble model.

In table~\ref{tab:inc-ta} we saw that, although POS tagging works for most of the cases, around 9\% of the ambiguities are only solvable by lemma prediction. Since it is possible to identify these instances during inference, an ensemble solution could use the lemma prediction model to disambiguate these.

Moreover, around 6\% of the instances currently cannot be disambiguated using either method.

This puts the upper limit on accuracy to 85\% for the better model (POS prediction). Using an ensemble model to also capture the lemma-only ambiguities would therefore push this limit to 94\%.

Another approach we have explored is the use of multi-task learning to predict both POS and lemma at the same time. We tried a na\"{\i}ve approach, reusing the LSTM parameters and alternating between the two different objectives during training. So far this has been somewhat unsuccessful, yielding an accuracy around 10\% lower than that of either of the single-task models, but we believe there is still much room for improvement.

\hyphenation{eith-er stand-alone}
To push the performance nearer to 100\%, it will be necessary to make a model that predicts morphological tags, either as an addition to the existing models, or as a standalone model that we can then invoke for these instances where the POS and lemma are the same.

\section*{Acknowledgements}

This work was supported in part by HIIT: Helsinki Institute for Information Technology.

\section*{Bibliographical References}
\label{main:ref}

\bibliographystyle{lrec}
\bibliography{./thebib-long,./revita}

\begin{thebibliography}{}

\bibitem[\protect\citename{Bojanowski \bgroup et al.\egroup }2017]{FastText}
Bojanowski, P., Grave, E., Joulin, A., and Mikolov, T.
\newblock (2017).
\newblock Enriching word vectors with subword information.
\newblock {\em Transactions of the Association for Computational Linguistics},
  5:135--146.

\bibitem[\protect\citename{Carreras \bgroup et al.\egroup }2004]{carreras04}
Carreras, X., Chao, I., Padr{\'o}, L., and Padr{\'o}, M.
\newblock (2004).
\newblock Freeling: An open-source suite of language analyzers.
\newblock In {\em Proceedings of the 4th International Conference on Language
  Resources and Evaluation (LREC'04)}.

\bibitem[\protect\citename{Dereza \bgroup et al.\egroup
  }2016]{dereza2016automatic-SOA-RUSSIAN}
Dereza, O., Kayutenko, D., and Fenogenova, A.
\newblock (2016).
\newblock Automatic morphological analysis for {Russian}: A comparative study.
\newblock In {\em Proceedings of the International Conference Dialogue}.

\bibitem[\protect\citename{Forcada \bgroup et al.\egroup }2011]{forcada2011afp}
Forcada, M.~L., {Ginest{\'i}-Rosell}, M., Nordfalk, J., {O’Regan}, J.,
  {Ortiz-Rojas}, S., {P{\'e}rez-Ortiz}, J.~A., {S{\'a}nchez-Mart{\'i}nez}, F.,
  {Ram{\'i}rez-S{\'a}nchez}, G., and Tyers, F.~M.
\newblock (2011).
\newblock Apertium: a free/open-source platform for rule-based machine
  translation.
\newblock {\em Machine Translation}, 25(2):127--144.

\bibitem[\protect\citename{Gers \bgroup et al.\egroup }2000]{gers2000}
Gers, F.~A., Schmidhuber, J.~A., and Cummins, F.~A.
\newblock (2000).
\newblock Learning to forget: Continual prediction with {LSTM}.
\newblock {\em Neural Computation}, 12(10):2451--2471.

\bibitem[\protect\citename{Hoya~Quecedo \bgroup et al.\egroup
  }2020]{2020-LREC-hoya-quecedo-disambig}
Hoya~Quecedo, J.~M., Koppatz, M.~W., Furlan, G., and Yangarber, R.
\newblock (2020).
\newblock Neural disambiguation of lemma and part of speech in morphologically
  rich languages.
\newblock In {\em Proceedings of LREC: the $12^{th}$ Conference on Language
  Resources and Evaluation}, pages 3573--3582. European Language Resources
  Association (ELRA).

\bibitem[\protect\citename{Inoue \bgroup et al.\egroup
  }2017]{inoue-etal-2017-joint}
Inoue, G., Shindo, H., and Matsumoto, Y.
\newblock (2017).
\newblock Joint prediction of morphosyntactic categories for fine-grained
  {A}rabic part-of-speech tagging exploiting tag dictionary information.
\newblock In {\em Proceedings of the 21st Conference on Computational Natural
  Language Learning ({C}o{NLL} 2017)}, pages 421--431, Vancouver, Canada.
  Association for Computational Linguistics.

\bibitem[\protect\citename{Kanerva \bgroup et al.\egroup }2018]{udst:turkunlp}
Kanerva, J., Ginter, F., Miekka, N., Leino, A., and Salakoski, T.
\newblock (2018).
\newblock Turku neural parser pipeline: An end-to-end system for the {CoNLL}
  2018 shared task.
\newblock In {\em Proceedings of the CoNLL 2018 Shared Task: Multilingual
  Parsing from Raw Text to Universal Dependencies}. Association for
  Computational Linguistics.

\bibitem[\protect\citename{Katinskaia and
  Yangarber}2018]{katinskaia:2018-DHN:revita}
Katinskaia, A. and Yangarber, R.
\newblock (2018).
\newblock Digital cultural heritage and revitalization of endangered
  {Finno-Ugric} languages.
\newblock In {\em Proceedings of the 3rd Conference on Digital Humanities in
  the Nordic Countries}, Helsinki, Finland.

\bibitem[\protect\citename{Katinskaia \bgroup et al.\egroup
  }2018]{katinskaia:2018-lrec:revita}
Katinskaia, A., Nouri, J., and Yangarber, R.
\newblock (2018).
\newblock Revita: a language-learning platform at the intersection of {ITS} and
  {CALL}.
\newblock In {\em Proceedings of LREC: 11th International Conference on
  Language Resources and Evaluation}, Miyazaki, Japan.

\bibitem[\protect\citename{Kingma and Ba}2015]{adam}
Kingma, D.~P. and Ba, J.
\newblock (2015).
\newblock Adam: A method for stochastic optimization.
\newblock {\em CoRR}, abs/1412.6980.

\bibitem[\protect\citename{Klyshinsky \bgroup et al.\egroup
  }2011]{klyshinsky2011method}
Klyshinsky, E., Kochetkova, N., Litvinov, M., and Maximov, V.
\newblock (2011).
\newblock Method of {POS}-disambiguation using information about words
  co-occurrence (for {Russian}).
\newblock {\em Proceedings of GSCL}, pages 191--195.

\bibitem[\protect\citename{Korobov}2015]{pymorphy}
Korobov, M.
\newblock (2015).
\newblock Morphological analyzer and generator for {Russian} and {Ukrainian}
  languages.
\newblock In Mikhail~Yu. Khachay, et~al., editors, {\em Analysis of Images,
  Social Networks and Texts}, pages 320--332, Cham. Springer International
  Publishing.

\bibitem[\protect\citename{Koskenniemi}1983]{koskenniemi:1983-thesis}
Koskenniemi, K.
\newblock (1983).
\newblock {\em Two-level morphology: A general computational model for
  word-form recognition and production}.
\newblock {Ph.D.} thesis, Helsinki, Finland.

\bibitem[\protect\citename{Kotelnikov \bgroup et al.\egroup
  }2017]{2017-kotelnikov-rus-morpho}
Kotelnikov, E., Razova, E., and Fishcheva, I.
\newblock (2017).
\newblock A close look at {Russian} morphological parsers: Which one is the
  best?
\newblock In {\em Conference on Artificial Intelligence and Natural Language},
  pages 131--142. Springer.

\bibitem[\protect\citename{Melamud \bgroup et al.\egroup }2016]{context2vec}
Melamud, O., Goldberger, J., and Dagan, I.
\newblock (2016).
\newblock context2vec: Learning generic context embedding with bidirectional
  {LSTM}.
\newblock In {\em Proceedings of The 20th SIGNLL Conference on Computational
  Natural Language Learning}, pages 51--61.

\bibitem[\protect\citename{Mikolov \bgroup et al.\egroup }2013]{word2vec}
Mikolov, T., Sutskever, I., Chen, K., Corrado, G., and Dean, J.
\newblock (2013).
\newblock Distributed representations of words and phrases and their
  compositionality.
\newblock In {\em Proceedings of the 26th International Conference on Neural
  Information Processing Systems}, volume~2, pages 3111--3119, USA. Curran
  Associates Inc.

\bibitem[\protect\citename{Moshagen \bgroup et al.\egroup
  }2013]{moshagen:2013:building-giellatekno}
Moshagen, S.~N., Pirinen, T., and Trosterud, T.
\newblock (2013).
\newblock Building an open-source development infrastructure for language
  technology projects.
\newblock In {\em Proceedings of NODALIDA 2013: the 19th Nordic Conference of
  Computational Linguistics}, pages 343--352, Oslo, Norway.

\bibitem[\protect\citename{Nivre \bgroup et al.\egroup }2007]{conll2007}
Nivre, J., Hall, J., K{\"u}bler, S., McDonald, R., Nilsson, J., Riedel, S., and
  Yuret, D.
\newblock (2007).
\newblock The {CoNLL} 2007 shared task on dependency parsing.
\newblock In {\em Proceedings of the 2007 Joint Conference on Empirical Methods
  in Natural Language Processing and Computational Natural Language Learning
  (EMNLP-CoNLL)}.

\bibitem[\protect\citename{Nivre \bgroup et al.\egroup }2018]{udtreebank}
Nivre, J., Abrams, M., and al.
\newblock (2018).
\newblock Universal dependencies 2.3.
\newblock {LINDAT}/{CLARIN} digital library at the Institute of Formal and
  Applied Linguistics ({{\'U}FAL}), Faculty of Mathematics and Physics, Charles
  University.

\bibitem[\protect\citename{Parra~Escart{\'i}n and
  Mart{\'i}nez~Alonso}2015]{2015-escartin-spanish}
Parra~Escart{\'i}n, C. and Mart{\'i}nez~Alonso, H.
\newblock (2015).
\newblock Choosing a {Spanish} part-of-speech tagger for a lexically sensitive
  task.
\newblock {\em Procesamiento del Lenguaje Natural}, 54:29--36.

\bibitem[\protect\citename{Schmid}2013]{2013:Schmid-probabilisticpart-of-speech-treetagger}
Schmid, H.
\newblock (2013).
\newblock Probabilistic part-of-speech tagging using decision trees.
\newblock In {\em New methods in language processing}.

\bibitem[\protect\citename{Tkachenko and
  Sirts}2018]{DBLP:journals/corr/abs-1810-06908}
Tkachenko, A. and Sirts, K.
\newblock (2018).
\newblock Neural morphological tagging for {Estonian}.
\newblock {\em CoRR}, abs/1810.06908.

\bibitem[\protect\citename{Wu \bgroup et al.\egroup }2016]{gtrans}
Wu, Y., Schuster, M., Chen, Z., Le, Q.~V., Norouzi, M., Macherey, W., Krikun,
  M., Cao, Y., Gao, Q., Macherey, K., Klingner, J., Shah, A., Johnson, M., Liu,
  X., Łukasz Kaiser, Gouws, S., Kato, Y., Kudo, T., Kazawa, H., Stevens, K.,
  Kurian, G., Patil, N., Wang, W., Young, C., Smith, J., Riesa, J., Rudnick,
  A., Vinyals, O., Corrado, G., Hughes, M., and Dean, J.
\newblock (2016).
\newblock Google's neural machine translation system: Bridging the gap between
  human and machine translation.
\newblock {\em CoRR}, abs/1609.08144.

\bibitem[\protect\citename{Yatbaz and Yuret}2009]{yatbaz:2009}
Yatbaz, M.~A. and Yuret, D.
\newblock (2009).
\newblock Unsupervised morphological disambiguation using statistical language
  models.
\newblock In {\em Proceedings of the NIPS Workshop on Grammar Induction,
  Representation of Language and Language Learning}, pages 321--324.

\bibitem[\protect\citename{Zalmout and Habash}2017]{zalmout}
Zalmout, N. and Habash, N.
\newblock (2017).
\newblock Don{'}t throw those morphological analyzers away just yet: Neural
  morphological disambiguation for {Arabic}.
\newblock In {\em Proceedings of EMNLP: Conference on Empirical Methods in
  Natural Language Processing}, pages 704--713, Copenhagen, Denmark.
  Association for Computational Linguistics.

\end{thebibliography}

\newpage
\clearpage
\appendix

\section{Examples of ambiguities}
\label{app:examples}

Additional examples of the kinds of ambiguities that our method handles (and does not
handle):

We divide surface form ambiguities into three categories in the following subsections: two
(or more) declinable lemmas, one declinable and one indeclinable lemma, or two indeclinable
lemmas.

We classify lemmas into two types---depending on whether they accept inflectional
morphemes: \emph{declinable} lemmas accept them, and \emph{indeclinable} lemmas do not.  Thus,
an indeclinable lemma has only one surface form.  Declinable lemmas can have many surface
forms.

In the examples, we use the following annotation convention:
\begin{itemizerCompact}
\item \surface{the surface form}
\item \lemma{lemma}
\item \translate{translation}
\item \morph{POS and morphological tags}
\end{itemizerCompact}

\subsection{Surface forms with two declinable lemmas}

{\bf Different lemma, different POS:} \\
Finnish surface form \surface{tuli} has two readings:\\

FI \surface{tuli}: \translate{fire $||$ s/he came} \\
\lemma{tuli} \translate{fire} \morph{Noun, nominative, sing.} $||$ \\
\lemma{tulla} \translate{come} \morph{Verb, indicative, active, \\
  \indent past tense, 3rd person, sing.}

\vspace{.5em}

\noindent
Russian surface form \surface{\textcyr{стали}} has two readings:\\

RU \surface{\textcyr{стали}}: \translate{steel $||$ they became}  \\
\lemma{\textcyr{сталь}} \translate{steel} \morph{Noun, genitive, sing.} $||$\\
\lemma{\textcyr{стать}} \translate{become} \morph{Verb, indicative, active, \\
  \indent past, 3rd person, plur.}

\vspace{.5em}

\noindent Spanish surface form \surface{vino} has two readings:\\

ES \surface{vino}: \translate{wine $||$ s/he came} \\
\lemma{vino} \translate{wine} \morph{Noun, sing.} $||$ \\
\lemma{venir} \translate{come} \morph{Verb, indicative, active, \\
  \indent past, 3rd person, sing.}

\vspace{.5em}
\noindent {\bf Same lemma, different POS:}
\vspace{.5em}

\noindent
Each of the following words (surface forms) has two readings, where the lemmas are the
same, but the POS are different:


ES \surface{parecer}{:} \\
\lemma{parecer} \translate{seem}, \morph{Verb, infinitive}  $||$ \\
\lemma{parecer} \translate{opinion}, \morph{Noun, sing.} \\

RU \surface{\textcyr{знать}}{:} \\
\lemma{\textcyr{знать}} \translate{know} \morph{Verb, infinitive} $||$ \\
\lemma{\textcyr{знать}} \translate{nobility} \morph{Noun, nominative, sing.}  \\

RU \surface{\textcyr{стать}}{:} \\
\lemma{\textcyr{стать}} \translate{become} \morph{Verb, infinitive} $||$ \\
\lemma{\textcyr{стать}} \translate{posture} \morph{Noun, nominative, sing.}  \\

RU \surface{\textcyr{печь}}{:} \\
\lemma{\textcyr{печь}} \translate{bake} \morph{Verb, infinitive} $||$ \\
\lemma{\textcyr{печь}} \translate{hearth} \morph{Noun, nominative, sing.} \\

\vspace{.5em}
\noindent {\bf Different  lemma, same POS:}
\vspace{.5em}

\noindent
This is type of ambiguity is present in all languages.  The following surface forms have
two (or more) readings:

FI \surface{palaa}{:} \\
\lemma{palaa} \translate{returns} \morph{Verb, present, 3rd, sing.} $||$ \\
\lemma{palata} \translate{burns} \morph{Verb, present, 3rd, sing.} \\

FI \surface{alusta}{:} \\
\lemma{alusta} \translate{pad, base} \morph{Noun, nominative, sing.} $||$ \\
\lemma{alus} \translate{ship} \morph{Noun, partitive, sing.} $||$ \\
\lemma{alunen} \translate{underlay} \morph{Noun, partitive, sing.} \\

RU \surface{\textcyr{черта}}{:} \translate{mark $||$ of the devil} \\
\lemma{\textcyr{черта}} \translate{mark} \morph{Noun, nominative, sing.} $||$ \\
\lemma{\textcyr{черт}} \translate{devil} \morph{Noun, genitive, sing.} \\

RU \surface{\textcyr{белку}}{:} \translate{squirrel (acc.) $||$ to the protein} \\
\lemma{\textcyr{белка}} \translate{squirrel} \morph{Noun, accusative, sing.} $||$ \\
\lemma{\textcyr{белок}} \translate{protein} \morph{Noun, dative, sing.} \\

ES \surface{\textcyr{fui}}{:} \translate{I was $||$ I went} \\
\lemma{\textcyr{ser}} \translate{be} \morph{Verb, past perf., 1st, sing.} $||$ \\
\lemma{\textcyr{ir}} \translate{go} \morph{Verb, past perf., 1st, sing.}

\vspace{.5em}
\noindent {\bf Same lemma, same POS:}
\vspace{.5em}

\noindent
These are the kinds of ambiguities that our methods do not address, since both the lemma
and POS are identical for the different analyses: \\

FI \surface{nostaa}{:} \\
\lemma{nostaa} \translate{raise}, \morph{Verb, infinitive}  $||$ \\
\noindent \lemma{nostaa} \translate{raise}, \morph{Verb, present, 3rd, sing.} \\


RU \surface{\textcyr{кота}}{:} \\
\lemma{\textcyr{кот}} \translate{cat} \morph{Noun, genitive, sing.} $||$ \\
\lemma{\textcyr{кот}} \translate{cat} \morph{Noun, accusative, sing.} \\

\subsection{Surface forms with one (or more) declinable and one indeclinable lemma}
\vspace{.5em}
\noindent {\bf Different lemma, different POS:}
\vspace{.5em}

ES \surface{sobre}{:} \\
\lemma{sobre} \translate{above} \morph{Preposition} $||$ \\
\lemma{sobre} \translate{envelope} \morph{Noun, sing.} $||$ \\
\lemma{sobrar} \translate{remain} \morph{Verb, present subjunctive, 1st/3rd, sing.} \\

RU \surface{\textcyr{уже}}{:} \\
\lemma{\textcyr{уже}} \translate{already} \morph{Adverb} $||$ \\
\lemma{\textcyr{уж}} \translate{grass snake} \morph{Noun, locative, sing.} $||$ \\
\lemma{\textcyr{узкий}} \translate{narrow} \morph{Adj, comparative}

p







\end{document}